\documentclass[a4paper,twoside]{article}

\usepackage{epsfig}
\usepackage{subcaption}
\usepackage{comment}
\usepackage{calc}
\usepackage{amssymb}
\usepackage{amstext}
\usepackage{amsmath}
\usepackage{amsthm}
\usepackage{multicol}
\usepackage{pslatex}
\usepackage{apalike}
\usepackage{titlesec}
\usepackage{SCITEPRESS}     

\titleformat{\paragraph}
{\normalfont\normalsize\bfseries}{\theparagraph}{1em}{}
\titlespacing*{\paragraph}
{0pt}{3.25ex plus 1ex minus .2ex}{1.5ex plus .2ex}
\begin{document}

\title{Modelling and Detection of Driver’s Fatigue using Ontology}

\author{
    \authorname{Alexandre Lambert\sup{1}, Manolo Dulva Hina \sup{1} Celine Barth\sup{1}, Assia Soukane\sup{1} and  \\ Amar Ramdane-Cherif\sup{2} }
    
   \affiliation{\sup{1}Inseec U Reasearch Center, ECE Paris School of Engineering, 37 quai de Grenelle, 75015 Paris, France}
    \affiliation{\sup{2}LISV Laboratory, Université de Versailles – Paris Saclay, 10-12 avenue de l’Europe, 78140 Vélizy, France}
    \email{alexandre.lambert@edu.ece.fr, {\{manolo-dulva-hina, celine.barth, assia.soukane\}}@ece.fr, rca@lisv.usvq.fr}
}


\abstract{Road accidents have become the eight leading cause of death all over the world. Lots of these accidents are due to a driver’s inattention or lack of focus, due to fatigue. Various factors cause driver’s fatigue. This paper considers all the measureable data that manifest driver’s fatigue, namely those manifested in the vehicle measureable data while driving as well as the driver’s physical and physiological data. Each of the three main factors are further subdivided into smaller details. For example, the vehicle’s data is composed of the values obtained from the steering wheel’s angle, yaw angle, the position on the lane, and the speed and acceleration of the vehicle while moving. Ontological knowledge and rules for driver fatigue detection are to be integrated into an intelligent system so that on the first sign of dangerous level of fatigue is detected, a warning notification is sent to the driver. This work is intended to contribute to safe road driving. }

\onecolumn \maketitle \normalsize \setcounter{footnote}{4} \vfill

\section{\uppercase{Introduction}}
\label{sec:introduction}

Road accidents have become the 8th cause of death worldwide. According to the European Road Safety Observatory \cite{european_road_safety_observatory_annual_2019} every year more than one million accidents take place in Europe, of which more than 25,000 are fatal. Each accident has a significant induced socio-economic cost for the country. More than a million people die on the roads around the world. While the trend is declining in Europe, the overall global trend is increasing as shown in the World Health Organization reports \cite{world_health_organization_world_2018}. 
One of the most fatal causes of these accidents is falling asleep on the wheel.US National Highway Traffic Safety Administration (NHTSA) estimates that more than 100,000 drivers are involved in fatigue-related accidents. Accidents caused by fatigue are among the most fatal causes of death \cite{salmon_bad_2019}. When we look at these same statistics for trucks accidents, more than 60$\%$ are related to fatigue. The use of expert systems for fatigue detection may be a good approach \cite{bishop_intelligent_2000}. 

By representing fatigue using a model, a level of fatigue can be detected and inferred continuously while reacting to a maximum number of situations so that the intelligent vehicle can ensure driver safety.
The work done to model fatigue did not use maximum parameters and focuses only on a small number of symptoms, as for the models proposed by \cite{bergasa_real-time_2006,friedrichs_camera-based_2010}. The results of these studies show that a representation of fatigue in the driving context helps to better ensure driver safety. However, this does not make these models totally reliable because they do not take into account all the factors that constitute fatigue. 

\raggedbottom Not all drivers react the same way to fatigue. In order to reproduce this disparity, studies such as \cite{liu_predicting_2009,thiffault_monotony_2003} modify parameters that affect drivers, such as time on task, sleep deprivation. 
To better represent fatigue in an intelligent vehicle, we proposed the use of various parameters from different sources to model the contexts of the vehicle, the driver, etc. and fusion them, taking into account the driver's profile.
The perception of the environment is essential in an intelligent system that detects driver’s fatigue, hence, the smart environment is equipped with sensors and web services that measure or deduce various parameters related to the context of the environment, the vehicle and the driver. In a smart city, these sensors and web services continuously retrieve relevant data. The obtained data are used to populate the fatigue model, hence, the model becomes a computer-readable representation of the driver's fatigue situation. Using data fusion and related rules, it is possible to determine the level of driver fatigue. Using the degree of driver’s fatigue, it is also possible to notify/alert the driver of an impending danger. 

\section{\uppercase{An intelligent system}}
Intelligent systems assist in meeting the needs of people, which are becoming increasingly complex. An intelligent system provides user with relevant and useful information, it is an intelligent collaborator, according to \cite{paul_jorion_principes_1989} vision. An intelligent system is usually connected to other systems via Internet of Things, and is able to provide a response suitable to the user’s need. \cite{kasabov_evolving_2006} stated that an intelligent system is capable of decision making, reasoning and action, and possesses knowledge. 

In our vision of an intelligent system, it should be: (1) able to process and react in real-time, (2) connected to other intelligent systems to be able to exchange information and be aware of a larger environment, (3) to interact with the outside world (engine, screen, etc.), (4) to sense the outside world through data (sensor data), (5) possesses knowledge about certain situations and must reason, (6) aware of possible errors that could come up, (7) able to adapt to its user by being aware of the user’s profile.

\begin{figure}[!h]
  \centering
   {\epsfig{file = 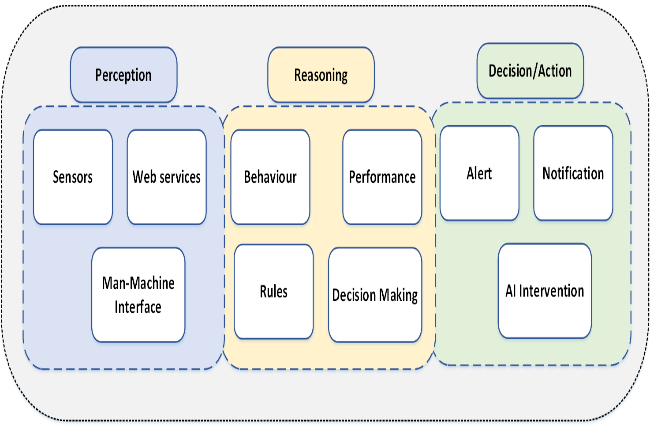, width = 7.5cm}}
  \caption{Basic components of an intelligent system.}
  \label{fig:basic_components_IS}
 \end{figure}

In the literature there are intelligent systems theories as proposed by \cite{albus_theory_1992,karray_soft_2004}. 
The intelligent system is composed of three components that interact with one other. The intelligent system components and the mechanisms inside each component depends on the type of application the intelligent system provides. The three components are common to intelligent systems. See Figure~\ref{fig:basic_components_IS}.

\subsection{The perception component}

The perception component allows interaction between the intelligent system and the outside world. Such interaction can be in sensing the environment using a sensor, a man-machine interface, or a web service. This component gathers all the information needed by the intelligent system. This module must be reliable. If it becomes unreliable, the data are transmitted to the system as well as the corresponding reasoning will be wrong. 

\subsection{The reasoning component}

The reasoning or "intelligent" component collects data from the perception component to determine the current situation and reasons out on this data and outputs a result induced from the input data. There are several ways to make a reasoning, but the goal remains the same: one set of input data yields a corresponding set of output data. Using appropriate rules, an intelligent system reasons out due to a system’s behaviour or performance. In this component, it is possible to use neural networks as a machine learning approach or inference engines coupled to a representation of ontological knowledge.

\subsection{The decision component}

Once the reasoning component has produced a result, it must be interpreted by the decision module in order to choose the action to be taken on the environment. The actions can be broad and depend on the type of intelligent system. But the goal remains the same, to impact an action on the environment according to the output data produced by the reasoning component. 

\section{\uppercase{An intelligent fatigue detection system}}
\begin{figure*}[!h]
  \centering
   {\epsfig{file = 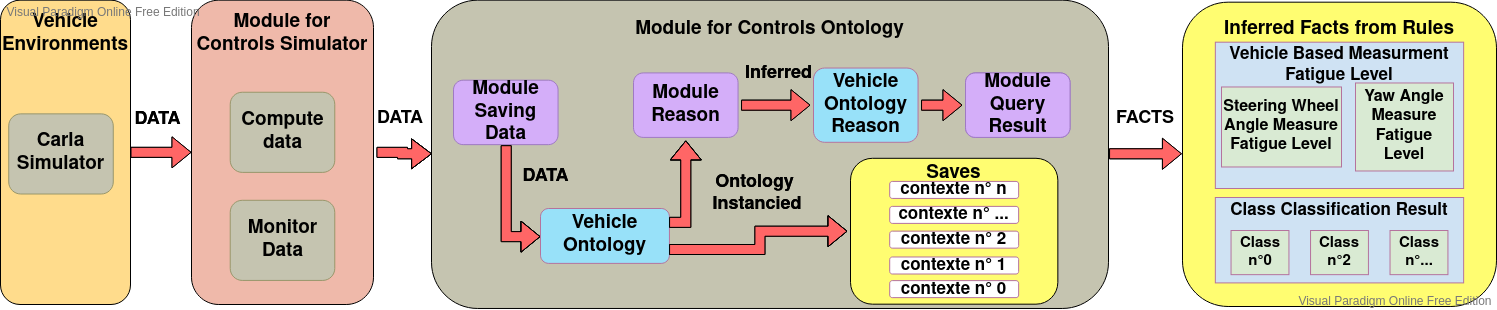, width = 15cm}}
  \caption{Schematic diagram of an intelligent driver fatigue detection system.}
  \label{fig:Schema_du_systeme}
 \end{figure*}

The schematic diagram of Figure~\ref{fig:Schema_du_systeme} describes a conceptual intelligent system for fatigue detection.  
The driving simulator is an environment in which the driving activity takes place. The driver interacts with the simulator and pilots the vehicle to move within such environment. As the vehicle moves, the simulator calculates parameters specific to the vehicle, such as speed, steering wheel angle, acceleration, etc.

The expert system is composed of two modules: the component that controls the data received from the simulator and the module that manipulates the ontology. 
Using the CARLA \cite{dosovitskiy_carla_2017} simulator, it is possible to retrieve data of certain parameters of interest, such as the steering angle, the yaw angle, or the acceleration. However, these measures must be recalculated to obtain new values, such as the average or the frequency. This is the functionality of the first component of the system which allows the acquisition of data from the simulator, its calculation and also the monitoring of the data in the form of logs. This module interrogates the simulator at a regular frequency and transmits calculated data to reasoning system.

The reasoning component is based on ontology, in which every data received will be inserted into the ontology as an instance or an attribute of a class. Here, using the actual data, one model ontology is transformed into an instantiated ontology. 
The reference to databases is still being tested, and will be integrated in our next publication. Hence, the data is saved into the ontology, which in itself will be a representation of the driver fatigue’s situation.
In order to keep a record of these contexts for later analysis, a version of the ontology is saved in a specific file, as a historical data. In parallel, the instantiated ontology is used by the reasoning module to classify and infer the class that indicates the level of driver’s fatigue. The reasoner will also use the rules in the ontology to merge different data.This will result in a reasoned ontology, which will have to be interrogated to find out the level of fatigue inferred.
\begin{figure}[!h]
  \centering
   {\epsfig{file = 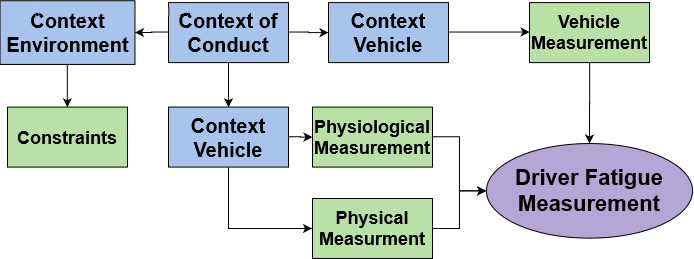, width = 7.5cm}}
  \caption{Driver fatigue model.}
  \label{fig:Composition_driver_fatigue_entiere}
 \end{figure}

It can thus be transmitted to the decision module which will take action according to this level. 
The last component in the diagram shows that the queried facts are inputs to the fatigue level measurement based on vehicle parameters (steering wheel angle, yaw angle measure). Based upon these values, a driver fatigue level classification is decided. 
As stated, a driver alert and notification mechanism follows to prevent further accident due to driver fatigue. This part, however, remains to be done and is part of future work.

\section{\uppercase{Driver fatigue model}}

Fatigue is a phenomenon characterised by a number of parameters that can be modelled
in terms of knowledge and that can manifest itself through an individual, such as a driver. This manifestation occurs through variations in the person’s physical and physiological states, as well as in his driving.(Figure~\ref{fig:Composition_driver_fatigue_entiere}) Fatigue manifests itself in the three broad categories of parameters: (1) Physical measurement, (2) Physiological measurement, and (3) Vehicle measurement. Each of these categories is broken down further into sub-categories and so on, down to the atomic parameters of the model. The purpose of the model is to list down all the parameters whose values may vary depending on the degree of one’s fatigue. It is thus possible to have values of the parameters that will be qualitative in terms of fatigue.

\subsection{Driver’s Fatigue Detection using Vehicular Measurements}

The vehicular measurements refer to the signals coming from the vehicle. These selected signals provide information on the driver's state of drowsiness, which have been studied by various researchers since 1990s \cite{dingus_development_1985,siegmund_correlation_1996}. 

\begin{figure}[!h]
  \centering
   {\epsfig{file =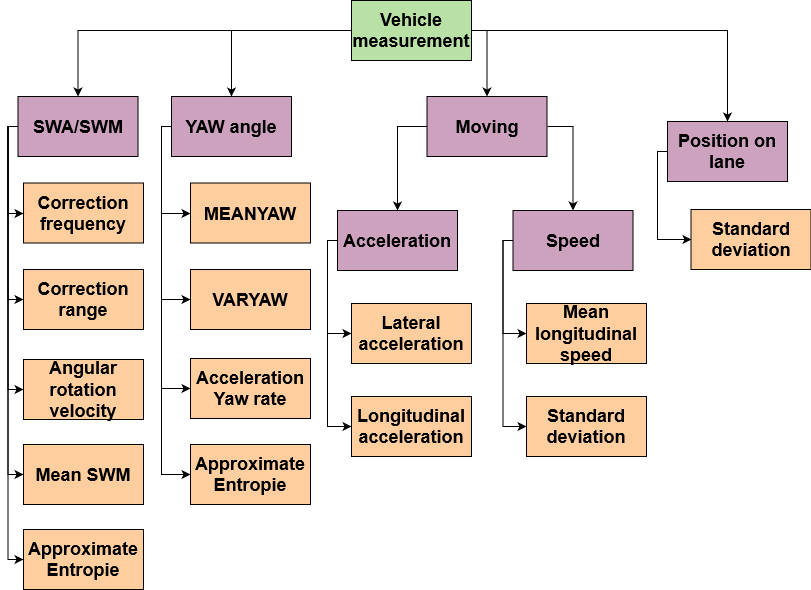, width = 7.5cm}}
  \caption{Vehicular parameters measurements.}
  \label{fig:Facteur_Mesure_Vehicule}
 \end{figure}
Studies on the impact of drowsy driving focus on one or more parameters \cite{li_support_2020,polychronopoulos_information_2004} but rarely use all measures and signals to describe fatigue. The driver fatigue detection using vehicular measurements (Figure~\ref{fig:Facteur_Mesure_Vehicule}). The model is based on five parameters, each of which is described below. This model is based on the results of previous studies that validated these parameters as presented by \cite{ramzan_survey_2019} in Table 1.

\subsubsection{Steering Wheel Angle (SWA/SWM)}

This measure is taken directly from the steering wheel angle sensor of the car, is expressed in degrees, and is a function of time. The variation of steering wheel angle over time may indicate information about driver drowsiness. The study in \cite{thiffault_monotony_2003} shows that a driver has a tendency to increase the amplitude of the steering wheel angle measurement during drowsy periods. 
During normal driving, the driver readjusts the car almost continuously with the help of micro-corrections that allow the car to stay in a lane. This work then led to several studies on the behaviour of the driver, notably in \cite{krajewski_steering_2009} which used the angle of the
steering wheel to deduce fatigue recognition patterns with a success rate of 86.1$\%$. 
During periods of fatigue, the driver is less sensitive to the small drifts of the car on the lane; this results in a correction of a higher amplitude with the steering wheel to refocus on the lane. \cite{thiffault_monotony_2003} estimated that the value of the angle over time for a driver was generally between 1$^{\circ}$ and 6$^{\circ}$, beyond which the value of the angle becomes significant in detecting a drowsy driver.
 It is also necessary to look at the time that a high angle (greater than 6$^{\circ}$) is maintained as this can provide information on the general movement of the car (a curved road, a bend, a large drift). It is necessary to take into account the shape and route of the journey in order to discern high angle values that are abnormal in a hypo vigilance case from normal. 
\paragraph{Steering wheel angle amplitude and  frequency}

High frequency of correction was significant for an alert driver because he was attentive to the position of the vehicle in the lane and wanted to keep it centred.
 Conversely, a low correction frequency may indicate a less alert driver because he is not aware of the vehicle's micro-drift and therefore tends to correct his trajectory less often as shown in \cite{ting_driver_2008}. The frequency of corrections can be described as normal, decreasing, or abnormally low depending on the number of steering wheel angle measurements greater than 6$^{\circ}$. The type of road should be taken into account to make the best use of this sub-parameter/factor. 

Amplitude of corrections describes the value that the steering wheel angle takes in degrees. Studies have shown that this value rarely exceeds 6° for an alert driver because the micro-correction helps in maintaining a trajectory is regular and does not require a "large steering wheel stroke" to correct deviations. Conversely, a driver in a fatigued state tends to have a large amplitude of steering wheel angle to correct deviations. In periods of fatigue, the driver is no longer aware of the drifts of his vehicle and realizes it later. This sub-parameter/factor is sensitive to road curvature, as a curved road may add an offset to the amplitude in order to maintain a trajectory.

\paragraph{Angular speed and average of the steering wheel}

\noindent This indicator describes the value of the angle over time in ($^{\circ}$/s). Studies have shown that a normal angular velocity value (less than 6$^{\circ}$/s) characterizes an alert and attentive driver while a high (greater than 6$^{\circ}$/s) and sustained angular velocity indicates the need to straighten the vehicle after a period of inattention. By analyzing the angular velocity in time windows, it is possible to notice the periods when the driver presents fatigue \cite{zhenhai_driver_2017}.
Average steering wheel angle describes the average value of the angle over a given period of time. It provides additional information and varies as the amplitude described above. It can be used to have a more general idea of the value of the angle in order to characterize the movement over time as opposed to the amplitude which describes the value of the angle at a given time. 
\paragraph{Operator type sub-parameters}
\noindent These sub-parameters/factors focus on transforming the signal to extract non-visible information. These operators can be of a mathematical nature or signal processing tools. Some studies use approximate entropy \cite{delgado-bonal_approximate_2019} and Fourier transforms to do this.

Approximate entropy is a mathematical operator which after a succession of operations allows a binary classification of the driver's state: tired or not. The approximate entropy is then extracted from the signal using adaptive piecewise linear approximation and classification.\cite{li_automatic_2017,li_online_2017} were able to detect the presence or absence of fatigue.

Frequency and phase function uses the furnace transformers to analyze the value of the angle as well as the angular velocity of the flywheel in frequency and phase. \cite{siegmund_correlation_1996} have shown that in the frequency domain, the power of the low frequency spectrum increases when the driver shows signs of fatigue. For the phase analysis, they expressed the phase of the steering wheel angle ($\theta$) as a function of the phase of the steering wheel angular velocity ($\omega$). An attentive and concentrated driver gives an expression of the phase concentrated in clusters around the origin ($\theta$ less than 6 $^{\circ}$, $\omega$ less than 25$^{\circ}$ per seconds) while a more tired driver will have an expression of the phase with larger loops ($\theta$ and $\omega$ larger) or small clusters at angles $\theta$ greater than 6 $^{\circ}$, showing inattentive behaviour typical of a tired driver. All these characteristics are analyzed using weighting functions.

\subsubsection{Yaw angle}

The yaw angle of a vehicle describes the rotational movements around the vertical axis of the vehicle. The yaw angle varies between 0° and 1° for an awake driver and above 1° for a sleeping driver.
The work of \cite{dingus_development_1985} shows the correlation between the mean and variance of the yaw angle and driver fatigue. \cite{li_automatic_2017} used yaw angle with approximate entropy to describe fatigue with good result. Also Yaw angle acceleration was compared for different driving situations including fatigue. A significant increase in yaw acceleration 2.5°/s\sup{2} was shown for a driver who was drowsy. The yaw angle sub-parameters include almost all of the steering wheel angle sub-parameters.

\subsubsection{Vehicle speed and acceleration}

\cite{chen_identification_2015} compared different acceleration rates (longitudinal and lateral), and noted a variation in the lateral acceleration rate of 2.0 m/s\sup{2} in the presence of fatigue. The results for the longitudinal acceleration are less encouraging but still worthy of inclusion in the model. It is always useful to re-study the parameter with our model

\subsubsection{The position on the track}

This sub-parameter describes the position of the vehicle on the track. As explained earlier, it is useful to focus on driver drift. One way to observe drift is to use the position on the sightline, including the standard deviation (deviation from a reference point) or to count the number of times the car crosses the line in the lane as studied by \cite{ting_driver_2008}. This indicator also makes it possible to observe whether the car is in a dangerous position for the driver and other cars (reversed direction, road overflow).

\subsection{Driver’s Fatigue Detection using Physical Parameters}
\begin{figure}[!h]
  \centering
   {\epsfig{file =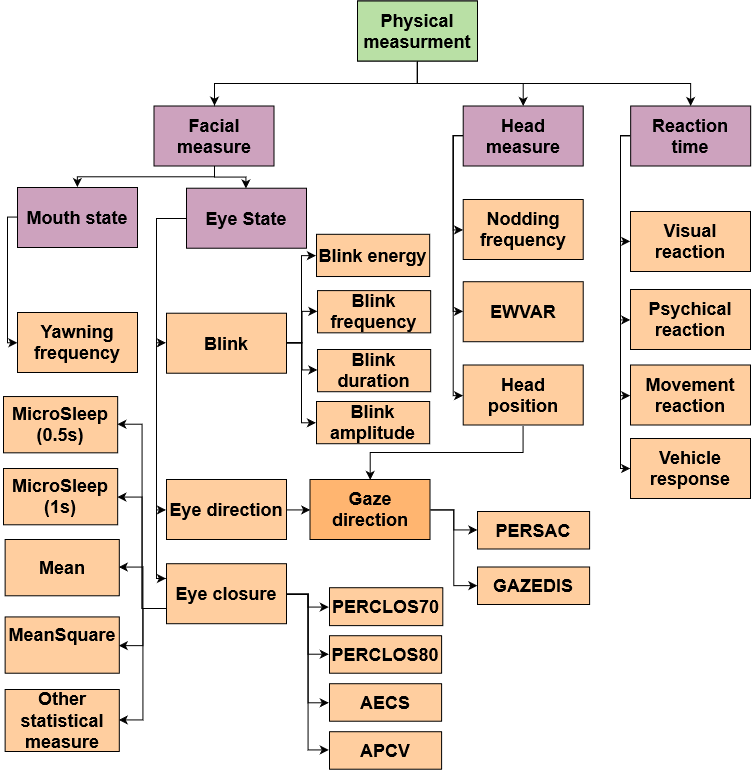, width = 7.5cm}}
  \caption{Physical parameters measurements.}
  \label{fig:Facteur_Comportement_Mesure}
 \end{figure}

This type of parameter groups together all relative changes in the driver's physical state. These measurements focus on the face, in particular on the eyes and mouth, which are very likely to change during periods of drowsiness (Figure~\ref{fig:Facteur_Comportement_Mesure}). 

\subsubsection{The Mouth’s Condition Parameter}

These parameters are very much studied by researchers because they are associated with facial recognition algorithms. They are among the most important because they contain decisive elements such as the eyes (e.g. closing the eye for a long time) to detect fatigue. The model consists of 4 main parameters, which describe the state of the mouth, the state of the eyes, the state of the head and the reaction time. Below, we have elaborated this sub-model.
Fatigue can manifest itself through the driver's yawning. It is commonly recognized as a sign of fatigue and should be taken into account in our model \cite{abtahi_driver_2011}. To do this, we studied the opening of the mouth, and if it exceeds a certain threshold for a significant time, then yawning occurs \cite{weiwei_liu_driver_2010,fan_yawning_2007}, or by calculating the air in the mouth \cite{reddy_real-time_2017}. \cite{azim_fully_2014} use the ratio and the air to qualify the state of the mouth. It is also possible to study the frequency of driver yawning in order to have an indication of the number of times the driver yawns, that if it increases may indicate significant fatigue. Other statistical operators (median, mean, etc.) can also be used to qualify the state of the mouth.

\subsubsection{Eye Condition Parameter}

There are 3 sub-parameters to be studied to deduce the state of the driver's eyes: the blinking, the opening of the eyes and the direction of the eyes. These parameters allow us to have a qualitative measure of the general condition of the driver's eyes. 

\paragraph{Blink Parameters}

\noindent As stated in \cite{schleicher_blinks_2008}, blinking and other associated measures are strongly related to fatigue, notably the duration of the blink which allows us to know how long the eye is occluded by the eyelid. This measure is useful for detecting driver micro sleep. Blink frequency is also commonly used because it has been shown that the number of blinks increases as driver alertness decreases \cite{hargutt_eyelid_2001}. Blink amplitude can help us to describe the state of the blink. In \cite{friedrichs_camera-based_2010}, all of these measures are used as well as the same measure but tailored to the driver.

\paragraph{Parameters related to eye closure}

\noindent In order to quantify eye closure, we can use an indicator that is widely recognized as effective, PERCLOS \cite{federal_highway_administration_perclos_1998}. This indicator represents the proportion of time the eye remains closed at least X\%\ over a time window. It is possible to use PERCLOS with larger or smaller time windows and a more or less significant percentage of eye closure. For example, \cite{friedrichs_camera-based_2010} uses 70\%\ PERCLOS and 80\%\ PERCLOS over a time window of 3 minutes. Other measurements are used such as square averages, average eye aperture, amplitude-velocity ration (APCV) or average eye velocity (AECS).
To properly qualify the condition of the eyes, the model must qualify the blink and the eye opening in frequency, amplitude and speed. Thus, it is possible to use other statistical operators (PERCLOS, mean square) to describe these indicators.

\subsubsection{Parameter Relating to Head Movement}

Head movements can also be an indicator of fatigue. Indeed, during periods of micro sleep that may occur at the wheel, the driver relaxes the muscles of his head \cite{hartley2000review}. Thus, it is possible to detect changes in head angle indicating a driver no longer has full control of his vehicle. This parameter is studied in frequency, in order to free oneself from head movements due to normal driving. In \cite{friedrichs_camera-based_2010}, they use a weighted exponential mean and variance (EWMA EWVAR).

\subsubsection{Eyesight Parameter}

The description of the movement of the head as well as the movement of the eye allow us to obtain the direction of the glance. In general, the direction of gaze can be used to describe several situations and not only fatigue. In particular, we can calculate the percentage of jerky eye movement (PERSAC) or the distribution of the gaze over time (GAZEDIS), which are parameters of the gaze related to fatigue \cite{ji_real-time_2004}.

\subsection{Driver reaction time}

This parameter describes different driver reaction times to obstacles (pedestrians, objects on the road). As stated in \cite{ting_driver_2008}, reaction times increase in the presence of driver fatigue. To qualify the driver's reaction time, 4-time intervals are defined: the visual reaction time of the obstacle, the physical reaction time, the movement time and the response time of the vehicle.

\subsection{Driver Fatigue Detection using Physiological Parameters}

\begin{figure}[!h]
  \centering
   {\epsfig{file =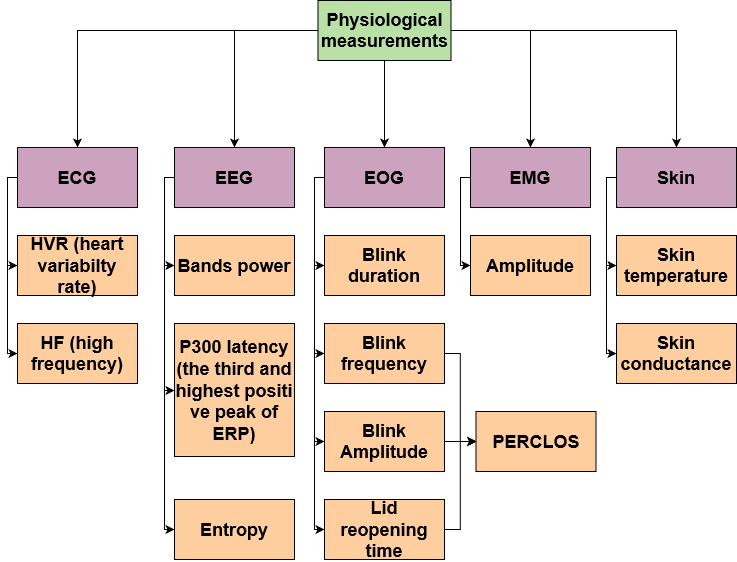, width = 7.5cm}}
  \caption{Physiological  parameters measurements.}
  \label{fig:Facteur_physiologique}
 \end{figure}

Physiological characteristics are very important, as they are robust, reliable and directly related to the physical and psychological state of the driver. Moreover, the acquisition is not directly disturbed by artefacts due to changes in weather or lighting conditions, unlike cameras as we have seen in the physical parameters \cite{begum_intelligent_2013}.
Measurements of physiological signals can be influenced by the general state (emotions and other psychological states) of the driver, so it is best to compare several of these measurements to ensure that the driver is fatigued. Some measures also vary from person to person. Physiological parameters can be divided according to the different modes of acquisition. There are parameters from the so-called "Electro" acquisition modes (EEG, ECG, EMG, EDA and EOG) and other parameters (Figure~\ref{fig:Facteur_physiologique}).

\subsubsection{Parameters from the EEGs}
The EEG parameters have been studied for quite a long time (the EEG of drowsiness in normal adults). The EEG allows the analysis of brain activity through different parameters \cite{chowdhury_sensor_2018}: the spectral power (e.g. delta, theta, alpha and beta bands), the amplitude and latency of the third peak of the ERP (Event Related Potential) as well as the entropic signal of the EEG.
After the transformation of the EEG signals, four EEG frequency bands are obtained. These four frequency bands are not precisely defined. However, the literature \cite{zhao_electroencephalogram_2012} frequently uses these amplitudes as having a frequency between approximately: Delta Band (1Hz-4Hz), Theta Band (4Hz-8Hz), Alpha band (8Hz-12Hz), Beta Band (12Hz-18Hz). The Beta band is significant as long as a cognitive task requires a high level of concentration (such as driving, for example). The Alpha band increases as alertness decreases (increases greatly during sleepy phases) but flattens after phase changes (e.g., alertness to drowsiness) \cite{brookhuis_monitoring_2010}. The Theta band finds its power in the primary phases of sleep. The Delta band is used to classify the states of intense sleep but it is almost null power during the sleeping states \cite{svensson_blink_2004}. Measurements of the different bands are likely to change from one subject to another. The acquisition of these signals in a vehicle can be complicated due to the noise and various artefacts.	

\subsubsection{ERP and EEG entropy}

An ERP is a response of the brain to a certain stimulus \cite{rondik_cognitive_2013}. The amplitude and latency of the third spade of the ERP is often studied. This information provides information about the cognitive resources and the speed of the driver's thinking time. Some studies have shown that the amplitude decreases with increasing driving time \cite{zhao_electroencephalogram_2012}. This implies a reduction in alertness and a longer reaction time.
Entropy is useful to qualify non-linear, unstable and dynamic signals such as EEGs. There are several types of entropies, for example \cite{mu_driver_2017} used four different entropies on an EEG data: Spectral Entropy, Approximate Entropy, Sample Entropy and Fuzzy Entropy. These parameters were used to train an SVM classification with EEG signals from 12 patients to classify fatigue. The results showed an algorithm performance of 98.75$\%$.

\subsubsection{Heart Rate and Its Variability (ECGs)}
ECG collects signals that provide information about an individual's cardiac system. These parameters correlate with drowsiness while driving. The heart rate is defined as the number of beats per minute (BPM). The reduction in heart rate can occur when the rider goes from awake to asleep as shown in \cite{sun_-vehicle_2011,furman_early_2008}.\cite{abdul_rahim_detecting_2015} uses a sensor placed on the steering wheel to monitor heart rate. BPMs are classified as normal between 75 and 100 for men and 70 and 95 for women and drowsy between 50 and 65 for men and 45 and 63 for women.
Heart rate variability which is defined as the variation in a time interval between two consecutive beats. It can be called R-R interval or RRI. The work of \cite{vicente_drowsiness_2016} tells us that the variation in heart rate can be described as the activity of the nervous system that is highly altered by fatigue.
HRV (heart rate variability) is also studied in spectral analysis, decomposing HRV into 3 bands \cite{shinar_autonomic_2006}: Very low frequency (0.0008Hz-0.04Hz), Low frequency (0.04Hz-0.15Hz), High frequency: (0.15Hz-0.5Hz). HRV activity in these frequency bands reveals the activity of the sympathetic nervous system \cite{chowdhury_sensor_2018}.

\subsubsection{Breathing Frequency}

Breathing frequency is the number of inhalations and exhalations per minute. \cite{sun_-vehicle_2011} shows a link between breathing rate and fatigue. However, other studies show no change in breathing cycles in the presence of fatigue \cite{shinar_autonomic_2006}. 

\subsubsection{Parameters from EMGs and EOG}

EMG is a signal generated by muscle contraction. The amplitude of the signal decreases as the driver begins to get tired \cite{chowdhury_sensor_2018}. However, signals from EMG are complex and random \cite{kumar_wavelet_2003}.The EOG signal is the potential of the electric field between the cornea and the retina, usually between 0.05 and 3.5 mV \cite{yue_eog_2011}. Each eye activity (blink, eye movement) changes this potential and thus the EOG signal \cite{thorslund_electrooculogram_2003}. Several measurements can be used to analyze the eye condition of the fatigued driver.A blink is defined as the contact of the upper and lower eyelid for at least 200 - 400 ms. The duration of the blink is therefore the time that this contact between the eyelids lasts \cite{yue_eog_2011}. If the blink duration is longer than 0.5 ms then it is a micro-sleep \cite{bando_evaluation_2017}. Blink Frequency is the number of blinks per minute, an increase in the number of blinks indicates a state of fatigue, as it is hard to keep your eyes open during this state. Blink Amplitude is the electrical potential measured during a blink. A normal blink varies between 100 - 400 µV. PERCLOS is the percentage of time in one minute when the eye is 80\%\ closed \cite{federal_highway_administration_perclos_1998}. Eyelid response time is the time it takes for the eyelid to reopen from the moment it is closed. This time increases as the driver gets tired.

\subsubsection{Skin temperature (ST) and Galvanic Skin Conductance (GSC)}
Studies have tried to show an association between skin temperature at certain body sites (nose, forehead) and fatigue levels as presented in \cite{bando_evaluation_2017}, however the results are too insignificant.
Galvanic Skin Conductance also known as EDA (Electro Dermal Activity), skin conductance is used more in the detection of stress than in the detection of fatigue as suggested by the study in \cite{rigas_reasoning-based_2008}. 

\section{\uppercase{Knowledge representation using ontology}}

According to \cite{gruber_translation_1993}, ontology is "an explicit specification of a conceptualization". \cite{studer_knowledge_1998} defined ontology as "an explicit and formal specification of a shared conceptualization", where "conceptualization" refers to an abstract model of a world phenomenon, and "explicit" means that the concepts used and the constraints on their use are explicitly defined. "Formal" means there is a rigorous technique for its specification and verification. "Shared" means that ontology captures knowledge that is consensual and accepted by a group. Among the various formalisms, we opted to edit our ontology using Protégé \cite{musen_protege_2015} in OWL/RDF format.
Among the various types of ontology proposed by \cite{gomez-perez_ontological_2004} (high-level, generic, domain, generic domain, etc.), our ontology contains the definitions to model a knowledge (fatigue).
Apart from being an application ontology type, our ontology also needs to be of the data and logic type \cite{roussey_ontologies_2010} below: 
Data ontologies provide a structural and syntactic description of domain concepts and their properties, where a concept is an aggregate of data to which it is possible to associate constraints from integrity on the values of these data.
Logical ontologies contain logical descriptions of concepts and relationships. Thy allow data integration from various sources. They contain logical formulas to be used by inference engines. These engines can validate an abstract model underlying a conceptualization to detect the class to which an instance belongs, or to recognize the instances of a class and generate new knowledge from rules.

\subsection{Driver Fatigue Representation}

Ontology is a good tool to represent contexts, and in particular the driving context. In this work, we consider a broader driving context. This driving context is described by \cite{hina_cognition_2018} as consisting of three parts: the vehicle, the environment and the driver. Together, they represent the driving context and make it possible to describe the various driving situations. In this work, however, we are interested not on a driving context but on the context of the driver's physiological state, fatigue. This context is composed of three sets: vehicle measurements, physical measurements, and physiological measurements. We represent driver fatigue in an ontological way through different measures described in the model. Ontological file have file extensions \textit{.owl, . rdf or .xml.} These files conform with the World Wild Web Consortium agreement which describes all the standards to adopt for the "semantic web".
Our ontology on driver’s fatigue is based on three main classes that describe our three concepts: \textit{vehicle measurements, physical measurements, and physiological measurements}

\subsection{Ontological Representation of Vehicular Measurements }
\begin{figure}\centering
{\epsfig{file =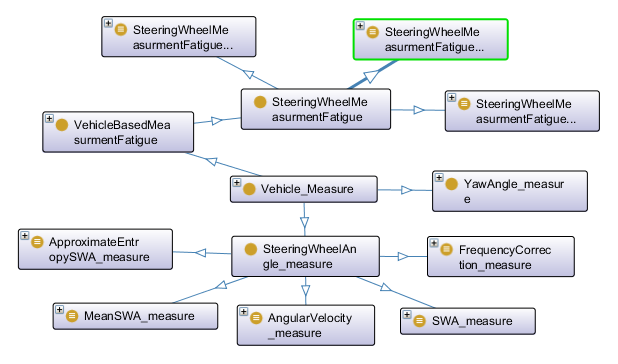, width = 7.5cm}}
\caption{Vehicle fatigue detection qualified with steering wheel angle measures (class representation)}
\label{fig:detection_vehicle_steer}
\end{figure}

\begin{figure}[!h]
\centering
{\epsfig{file =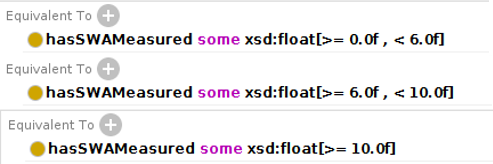, width = 7.5cm}}
\caption{Equivalence to qualify SWA class}
\label{equivalence_to_qualify_SWA}
\end{figure}

The \textit{Vehicle$\_$Measure} class describes all the vehicle measurements of our model. It is divided into 3 sub-classes: \textit{SteeringWheelAngleMeasurement, YawAngleMeasurement, VehicleBasedMeasurementFatigue}. The \textit{SteeringWheelAngleMeasurment} is used to describe the various measurements derived from the angle of a steering wheel. It is divided into 5 sub-classes which are: \textit{MeanSWA, FrequencySWA, SWA$\_$measure, ApproximateEntropySWA, AngularVelocity} (Figure~\ref{fig:detection_vehicle_steer})
All these sub-classes represent measurements from sensors or measurements calculated from sensor data (such as frequency, mean or variance). These classes have \textit{"DataProperties"} of type \textit{"hasSomeValue"} which represents the value of the measurement.The data property \textit{"hasSWAMeasured"} describes a relation of membership of a float type value to the \textit{SWA$\_$measure} class. The data property has for domain (or intersection) the \textit{SWA$\_$measure} class and for range a float type. The same relationships are applicable for the other classes. To qualify these measures, the five sub-classes are divided further into three (two for \textit{AngularVelocity}) sub-classes. 
As an example, \textit{SWA$\_$measure} is divided into \textit{SWA$\_$Extreme (more than 10$^{\circ}$), SWALarge (from 10$^{\circ}$ to 6$^{\circ}$), SWA$\_$Small (from 6$^{\circ}$ to 0$^{\circ}$)} (Figure~\ref{equivalence_to_qualify_SWA})

\begin{figure}[!h]
\centering
{\epsfig{file =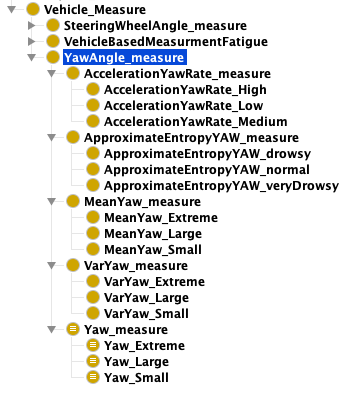, width = 7.5cm, height = 7.5cm}}
\caption{Equivalence to qualify SWA class}
\label{hierachyYawQualifing}
\end{figure}

The \textit{YawAngle$\_$Measure} describes the different measurements resulting from the yaw angle. It is divided into 5 sub-classes, namely: \textit{MeanYaw, VarYaw, Yaw, ApproximateEntropyYaw,} and \textit{AngularVelocityYaw}.
All these sub-classes have \textit{DataProperties} of type \textit{hasSomeValue} which represents the value of the measurement. 
The data property \textit{hasYawAngleMeasured} of float value related to the class \textit{Yaw$\_$measure}.
The same relationships exist for other classes. To quantify each of the five measures, each measure is divided into three sub-classes. For example, \textit{Yaw$\_$measure} is divided into \textit{Yaw$\_$Extreme, Yaw$\_$large, Yaw$\_$Small}, signifying the three types of quantifying the Yaw measurement.
The same qualification of measurements is applied to the sub-classes describing steering wheel angle, resulting in a class qualification as shown in the class hierarchy  (Figure~\ref{hierachyYawQualifing}).

The \textit{VehicleBasedMeasurementFatigue} class infers a fatigue level using the vehicle parameters. This qualification is different from the other classes being a so-called "knowledge" class. This class reports on the state of qualification of other classes at a given instant \textit{t}. 

\subsection{Ontological Representation of Physical Measurements}

\begin{figure}[!h]
    \centering
   {\epsfig{file =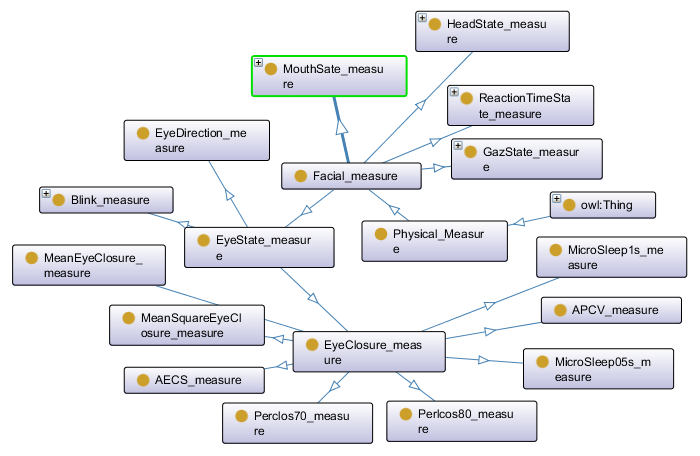, width = 7.5cm}}
\caption{Sub-classes derived from facial measurements with d classes describing eye closure}
\label{fig:Sub_classes_facial_eye_closure}
\end{figure}

The \textit{Physical$\_$measure} class describes the physical measurements related to a driver’s fatigue. It has a sub-class, \textit{Facial$\_$measure}, which describes the driver's facial measurements. This measurement is divided into different sub-classes that represent regions of the face (eyes, mouths, head). Each of these classes has its own sub-classes, further describing the class. For example, the eye closure measurements.In the same way as the vehicle parameters, the physical parameters have data properties that allow each class to describe a relationship of a measurement value from the sensors or from a calculation. (Figure~\ref{fig:Sub_classes_facial_eye_closure})

\begin{table*}[h]
\caption{Various vehicular parameters combination and the resulting driver fatigue level.}
\label{tab:rules_driver_fatigue}
\resizebox{\textwidth}{!}{%
\centering
\begin{tabular}{|l|l|l|l|l|} 
\hline
\multicolumn{4}{|l|}{Inputs}                                                            & Output                                           \\ 
\hline
MeanSWA$\_$Small   & AngularVelocity$\_$Normal & FrequencyCorrection$\_$Low    & SWA$\_$Small   & \textbf{SteeringWheelMeasurmentFatigue$\_$Low}     \\ 
\hline
MeanSWA$\_$Large   & AngularVelocity$\_$High   & FrequencyCorrection$\_$Normal & SWA$\_$Large   & \textbf{SteeringWheelMeasurmentFatigue$\_$Medium}  \\ 
\hline
MeanSWA$\_$Extreme & AngularVelocity$\_$High   & FrequencyCorrection$\_$High   & SWA$\_$Extreme & \textbf{SteeringWheelMeasurmentFatigue$\_$High}    \\ 
\hline
MeanSWA$\_$Large   & VarYaw$\_$Large           & AccelerationYawRate$\_$Medium & Yaw$\_$Large   & \textbf{YawAngleMeasurmentFatigue$\_$Medium}       \\ 
\hline
MeanSWA$\_$Small   & VarYaw\_Small           & AccelerationYawRate$\_$Low    & Yaw$\_$Small   & \textbf{YawAngleMeasurmentFatigue$\_$Low}          \\ 
\hline
MeanSWA$\_$Small   & VarYaw$\_$Extreme         & AccelerationYawRate$\_$High   & Yaw$\_$Extreme & \textbf{YawAngleMeasurmentFatigue$\_$High}         \\
\hline
\end{tabular}}
\end{table*}
\subsection{Ontological Representation of Physiological Measurements}
The representation of the ontology of physiological measurements follows the same approach as the other representations (see Section 5.2 and 5.3)
\subsection{Rules for Detecting and Reasoning on Driving Fatigue}
Once the general fatigue model for qualifying the data has been implemented in the ontology, rules must be established to standardize our ontology. It is on these rules that the level of fatigue that will be inferred is based. The rules use instances of classes or individuals in order to operate the reasoning.
SWRL (Semantic Web Rule Language) is the language used to express first order rules. The majority of rules are of the type: 
If \textit{"an instance of class A" and "an instance of the class X1"} and \textit{"has an attribute Y1"} and \textit{"Y1 satisfies such conditions"} and \textit{"an instance of the class X2"} and \textit{"has an attribute Y2"} and \textit{"Y2 satisfies such conditions"} then \textit{"the instance of class A belongs to class B"}. 
 
 Written rules are not based on precise values but on fuzzy values, this is based on the concept of fuzzy logic [46]. Indeed, parameter values are qualified according to fuzzy notions such as: \textit{strong, very strong, a little bit, little, etc.} which use these fuzzy values to infer knowledge from the written rules.
In our case, the type of rules on fatigue level are simpler because the instances of the classes of the model parameters have already been qualified.The rules that are used in ontology are as follows:  
If \textit{"there is an instance of the \textit{SteeringWheelMeasurementFatigue} class"} and \textit{"there is an instance of the \textit{SWA$\_$Extreme} class"} and \textit{"there is an instance of the \textit{AngularVelocity$\_$High} class"} and \textit{"there is an instance of the \textit{MeanSWA$\_$Extreme} class"} and \textit{"there is an instance of the \textit{FrequencyCorrection$\_$High} class"} then \textit{"the instance of the \textit{SteeringWheelMeasurementFatigue} class belongs to the \textit{SteeringWheelMeasurementFatigue$\_$High} class"}.
Table~\ref{tab:rules_driver_fatigue} summarizes the rules, implemented in the ontology, to infer a level of fatigue from measurements of steering wheel angle and yaw angle. For each Input (Qualified Class), an Output  is associated with it (Qualitative fatigue measurement from steering wheel angle and yaw angle).  
By writing rules like this, one can associate qualified classes with a level of fatigue. To limit the number of rules, it makes sense not to merge all the parameter but to have sub-merge that allows intermediate fatigue states to be reported. The principle of sub-merge shows that it is possible to efficiently reduce the number of rules to be written by merging a large set of data. Currently, there are 6 rules in Table~\ref{tab:rules_driver_fatigue}. We can generate more rules but we are still conducting experiments. A larger number of rules and the results of these experiments will appear in our next publication.

\subsection{Prospect for Data Fusion}
To use our sets of rules in an optimal way, it is possible to use  a weight system, so that some parameters are more valuable than others. This way, the determining parameters will have a greater importance in detecting fatigue or use a neural network so that the inference engine (reasoner) uses only the rules that are useful for a given context (the neural network would be trained using the different contexts that are saved by an intelligent system).

\section{\uppercase{Use case simulation}}

\begin{figure}[!h]
  \centering
   {\epsfig{file = 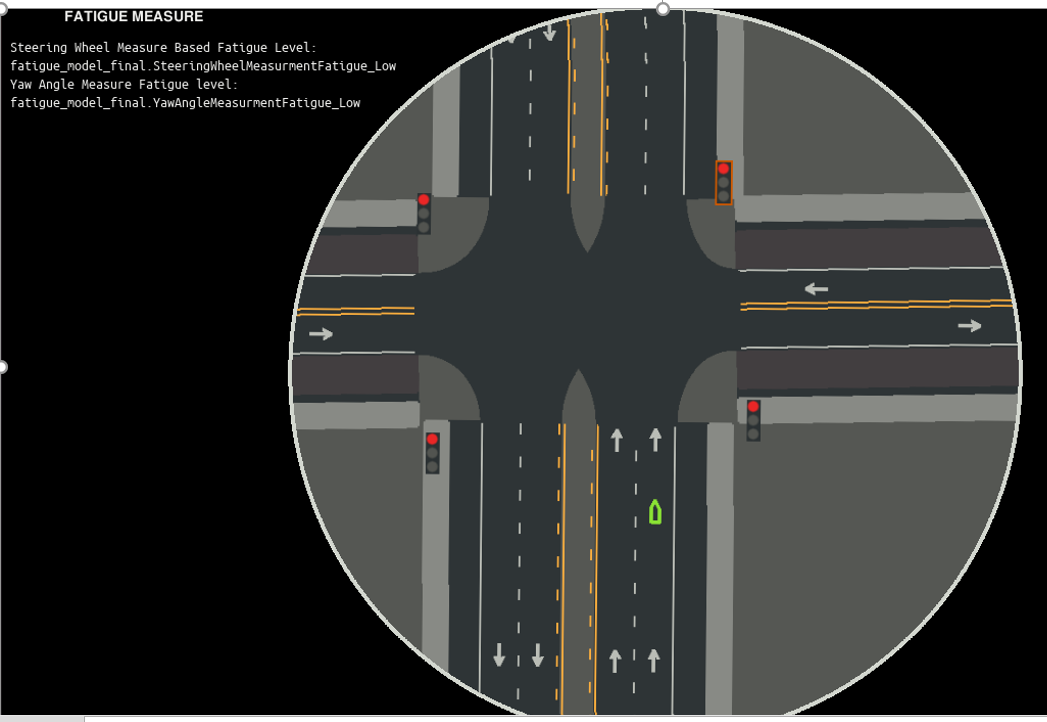, width = 7.5cm}}
  \caption{Carla simulator with the expert system fatigue.}
  \label{fig:carla_simulator}
 \end{figure}

In order to test our fatigue model, and more largely our intelligent system, we need data that can be used by the system. For simplicity we use a driving simulator to generate data.  Carla Simulator, is an open-source driving simulator for research in autonomous car \cite{dosovitskiy_carla_2017}. This simulator offers a large choice of modularity in the selection of driving environments, climatic conditions, circuit choices, etc. It is also possible to simulate vehicle data such as steering wheel angle, yaw angle and angular velocity. These parameters are present in our model and are used to calculate other sub-parameters. 
In order to create a use case, we need to integrate our intelligent system which is composed of different modules described in Figure~\ref{fig:Schema_du_systeme} (data extraction and computation, ontology control module and knowledge retrieval module). This allows our intelligent system to retrieve the data and process it in parallel with the use of the simulator. Thus, we can display the newly created information that serves as an alert to the driver (Figure~\ref{fig:carla_simulator}). As shown, two parameters are fused. The \textit{"SteeringWheelMeasurementFatigue"} is derived from the steering wheel angle while \textit{"YawAngleMeasurementFatigue"} is derived from the yaw angle. The mode used in this exercise, however, does not allow handling the vehicle with precisions unlike the normal mode. This is due to the technical constraints of the computer. Hence, the integration of the intelligent system in the simulator is a future work so that it will become a more appropriate simulation tool with more appropriate resources and capacity.
\section{\uppercase{Conclusions}}
\label{sec:conclusion}
In this paper, three factors, namely the vehicular data as well as the driver’s physical and physiological data are considered and combined to determine the level of driver’s fatigue. Each of these three factors are further subdivided into smaller measurable data, each of which becomes a contributing factor in the global assessment of a driver’s fatigue. A simple use case scenario is tested to detect  driver’s fatigue.  The research is ongoing, and future work is ongoing to determine the mathematical or computational method of combining them. The main idea is to supervise and measure each of these factors in order to perceive the environment and the level of driver’s fatigue. Ontology is used for knowledge representation and SWRL rules are used to compute for the level or degree of driver’s fatigue. In case of a detection of a dangerous level of fatigue, a notification is sent to the driver. Altogether, this methodology will be integrated into an intelligent system that is capable of detecting driver’s fatigue and in general, contribute to safe road driving.  
\bibliographystyle{apalike}
{\small
\bibliography{driver_fatigue.bib}}
\end{document}